\documentclass[letterpaper, 10pt, conference]{ieeeconf}
\IEEEoverridecommandlockouts
\usepackage{times}
\usepackage{graphicx}
\usepackage{amsmath}
\usepackage{amssymb}
\usepackage{booktabs}
\usepackage{multirow}
\usepackage{xcolor}
\usepackage{url}
\usepackage{subcaption}
\usepackage{microtype}
\usepackage[hidelinks]{hyperref}
\usepackage{balance}

\graphicspath{{figs/}}

\newcommand{\eg}{\textit{e.g.}}
\newcommand{\etal}{\textit{et al.}}

\title{\LARGE \bf
Severe Domain Shift in Skeleton-Based Action Recognition:\\
A Study of Uncertainty Failure in Real-World Gym Environments}

\author{\authorblockN{Aaditya Khanal\authorrefmark{1} and Junxiu Zhou\authorrefmark{1}}
\authorblockA{\authorrefmark{1}School of Computing and Analytics,\\
Northern Kentucky University, Highland Heights, KY 41099, USA\\
\{khanala1, zhouj2\}@nku.edu}}

\begin{document}
\maketitle
\thispagestyle{empty}
\pagestyle{empty}

\begin{abstract}
The practical deployment gap---transitioning from controlled multi-view 3D skeleton capture to unconstrained monocular 2D pose estimation---introduces a compound domain shift whose safety implications remain critically underexplored. We present a systematic study of this \textit{severe domain shift} using a novel Gym2D dataset (style/viewpoint shift) and the UCF101 dataset (semantic shift). Our Skeleton Transformer achieves $63.2\%$ cross-subject accuracy on NTU-120 but drops to $1.6\%$ under zero-shot transfer to the Gym domain and $1.16\%$ on UCF101. Critically, we demonstrate that high Out-Of-Distribution (OOD) detection AUROC does not guarantee safe selective classification. Standard uncertainty methods fail to detect this performance drop: the model remains \textit{confidently incorrect} with $99.6\%$ risk even at $50\%$ coverage across both OOD datasets. While energy-based scoring (AUROC $\ge 0.91$) and Mahalanobis distance provide reliable distributional detection signals, such high AUROC scores coexist with poor risk-coverage behavior when making decisions. A lightweight finetuned gating mechanism restores calibration and enables graceful abstention, substantially reducing the rate of confident wrong predictions. Our work challenges standard deployment assumptions, providing a principled safety analysis of both semantic and geometric skeleton recognition deployment.
\end{abstract}

\section{INTRODUCTION}

Skeleton-based action recognition powers emerging applications from rehabilitation robotics \cite{stgcn} to AI-powered personal coaching. While large benchmarks like NTU RGB+D 120 \cite{ntu120} drive rapid progress, deployment to real-world gym environments introduces compounding challenges that invalidate standard safety assumptions.

The domain gap has two distinct components. First, a \textbf{modality gap}: benchmark datasets provide precise 3D MoCap skeleton joints, while real-world deployment relies on monocular 2D pose estimation. Second, an \textbf{environment gap}: studio-quality multi-camera recording vs.\ uncontrolled single-view gym video. Together, these gaps cause severe accuracy degradation.

Crucially, standard uncertainty quantification (UQ) methods fail to detect this degradation. In a safety-critical fitness context, this failure is significant: a system that is confidently incorrect may provide injurious feedback to users executing loaded compound movements (\eg, barbell squats). We expose a critical flaw in current evaluation paradigms: high OOD detection AUROC does \textit{not} translate to safe selective classification.

We make the following central contributions:
\begin{itemize}
  \item \textbf{Safety Insight:} We demonstrate that high OOD AUROC can coexist with poor selective classification risk-coverage, challenging the assumption that distributional separation implies safe deployment.
  \item \textbf{Deployment Degradation:} We isolate and quantify severe zero-shot domain shift, showing how standard UQ methods fail to discriminate confident correct predictions from confident errors.
  \item \textbf{Gym2D Dataset:} We introduce a practical dataset derived from real-world gym videos to study these deployment gaps.
  \item \textbf{Graceful Abstention:} We present a finetuned gating mechanism that transitions an unsafe model into one exhibiting graceful abstention, substantially reducing the Wrong-Spoke Rate.
  \item \textbf{Failure Mode Analysis:} We identify that semantic confusion (\eg, visual similarity between squat and deadlift) and geometric projection drive systematic confident mis-classification, validated across multiple architectures (Skeleton Transformer and ST-GCN \cite{stgcn}) and Deep Ensembles.
\end{itemize}

\begin{figure}[t]
  \centering
  \includegraphics[width=\linewidth]{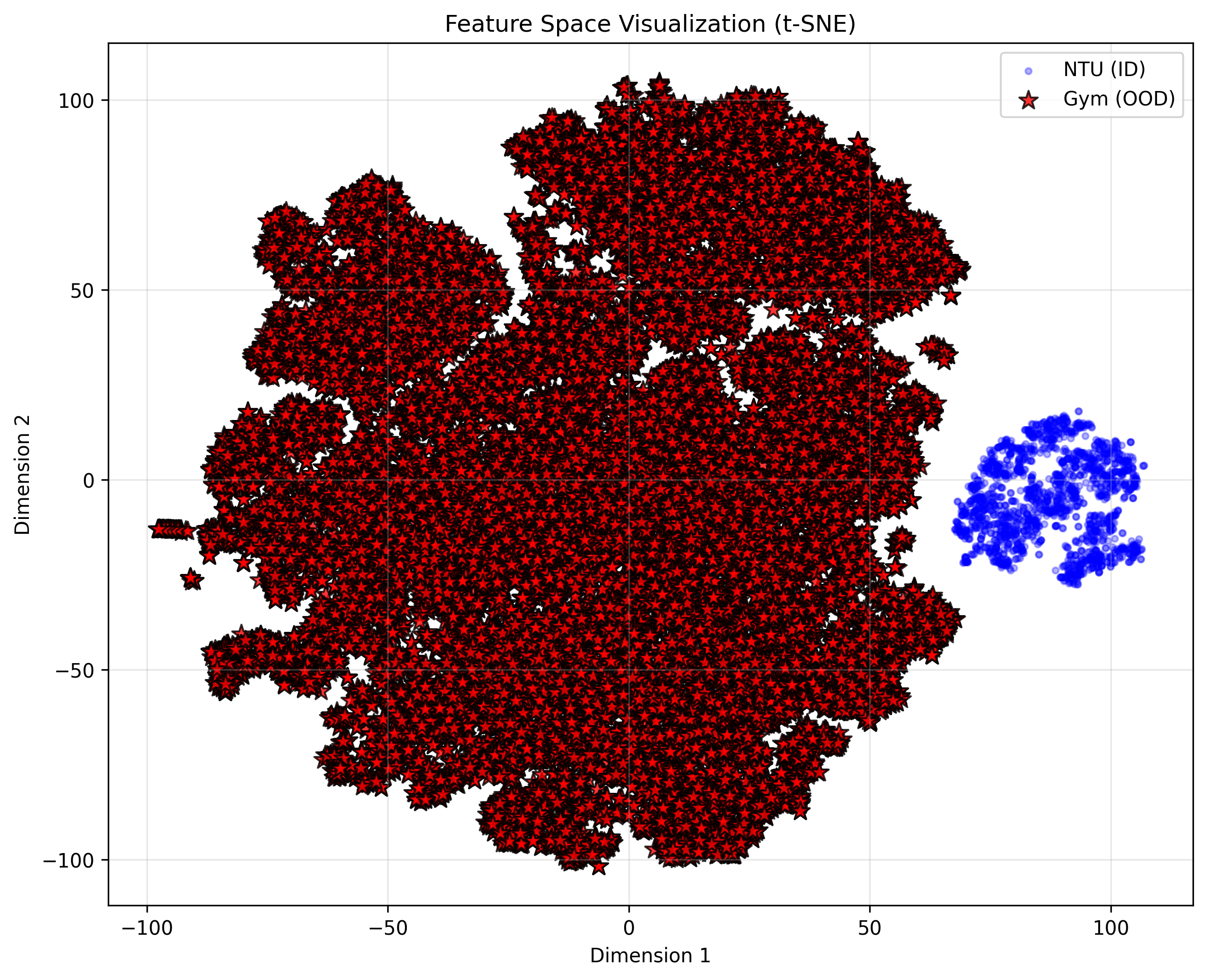}
  \caption{t-SNE visualization of Skeleton Transformer features. NTU-120 (ID) classes form well-separated clusters, while Gym2D (OOD) samples (red) collapse into the interior of NTU clusters despite being systematically mis-classified. This explains why softmax confidence remains high for wrong predictions.}
  \label{fig:tsne}
\end{figure}

\section{RELATED WORK}

\subsection{Skeleton-Based Action Recognition}
ST-GCN \cite{stgcn} introduced spatial-temporal graph convolutions, establishing the dominant paradigm. Subsequent work (CTR-GCN \cite{ctrgcn}, InfoGCN \cite{infogcn}, PoseConv3D \cite{poseconv3d}) improves accuracy via topology refinement, contrastive objectives, and 3D heatmap volumes. Transformer-based methods \cite{dstanet,skeletr} capture long-range dependencies. All are evaluated exclusively on in-distribution benchmarks; we study what happens under real-world OOD deployment.

\subsection{Uncertainty Quantification and OOD Detection}
MC Dropout \cite{mc_dropout} and Deep Ensembles \cite{deep_ensembles} estimate epistemic uncertainty; Temperature Scaling \cite{temp_scaling} provides post-hoc calibration. Ovadia et al.\ \cite{ovadia2019} showed these methods degrade under distribution shift---we verify and quantify this failure for skeleton recognition. Energy scoring \cite{energy_ood} uses the log-sum-exp of logits as an OOD signal; Mahalanobis distance \cite{mahalanobis_ood} measures Gaussian class-conditional distances in feature space. We are the first to systematically compare these methods for skeleton recognition under real-world compound domain shift and expose their selective classification failure.

\subsection{Test-Time Adaptation}
TTA methods like TENT \cite{wang2020tent} minimize entropy on unsupervised targets during deployment. Under severe structural domain shift, TTA often suffers from confirmation bias---reinforcing confidently wrong predictions rather than correcting them, because entropy minimization drives the model to become even more certain about whatever incorrect class it initially predicts. Our safety paradigm relies instead on explicit supervised finetuning for reliable calibration. Conformal prediction \cite{conformal} provides distribution-free coverage guarantees, but requires calibration data from the target domain and does not address the underlying confidence pathology we expose.

\section{PROBLEM FORMULATION \& DATASETS}

\subsection{Datasets and Domain Gap}
Let $\mathcal{D}_{S}$ be the NTU RGB+D 120 source domain \cite{ntu120}: 3D skeleton sequences $\mathbf{x} \in \mathbb{R}^{T \times 25 \times 3}$ ($T{=}60$ frames, $C{=}120$ classes, cross-subject split). We evaluate on two OOD target domains:

\noindent\textbf{1. Gym2D (Style/Geometric Shift):} 29{,}005 exercise sequences $\mathbf{x} \in \mathbb{R}^{T \times 17 \times 2}$ (COCO 17-keypoint via RTMPose \cite{rtmpose}, 99 exercise categories) extracted from real-world gym footage with diverse lighting, clutter, anthropometry, and camera angles. Monocular 2D estimation introduces depth ambiguity and self-occlusion under extreme joint angles.

\noindent\textbf{2. UCF101 (Semantic/Structural Shift):} 13{,}320 sequences (101 action categories) representing far-OOD semantic shift. Pose coordinates extracted via 2D estimation provide a vocabulary shift distinct from Gym2D's near-OOD geometric domain. UCF101 also captures a broader demographic and environmental diversity than studio-recorded benchmarks, further amplifying the distributional distance from NTU-120.

\subsection{Topological Modality Mapping (2D $\to$ 3D)}
To evaluate 2D targets against a 3D backbone, we map the $17$ COCO keypoints to their anatomical $25$-joint counterparts. The $(x,y)$ channels map directly; the $z$-depth channel and non-overlapping joints are zero-padded, preserving the $T \times 25 \times 3$ tensor format while manifesting the full geometric severity of the modality gap.

\subsection{Selective Classification and Safety}
We adopt the Selective Classification framework \cite{selective_clf}. A model accepts fraction $\kappa$ (coverage) of inputs with Risk $R(\kappa)$:
\begin{equation}
  R(\kappa) = \frac{\sum_{i \in \mathcal{A}_\kappa} \mathbf{1}[\hat{y}_i \neq y_i]}{|\mathcal{A}_\kappa|}
\end{equation}
where $\mathcal{A}_\kappa$ retains the $\kappa$-fraction of most confident samples. We focus on the \textit{Wrong-Spoke Rate} (WSR) $= R(\kappa) \times \kappa$---the fraction of all inputs where the model confidently predicts the wrong class. In fitness contexts, abstaining causes no harm while a confident wrong prediction on heavy compound movements directly increases injury risk. A safe coaching system must strictly minimize WSR.

\section{MODELS AND METHODS}

\subsection{Skeleton Transformer}
Joint-level tokenization with positional embeddings $d_j{=}32$, projected to $d_{\text{model}}{=}256$, with $L{=}4$ transformer encoder layers and multi-head self-attention over $T{\cdot}J$ tokens. Trained for $80$ epochs with Adam ($\text{lr}{=}10^{-3}$, weight decay $10^{-4}$), achieving $63.2\%$ NTU cross-subject accuracy (mean over 3 seeds).

\subsection{ST-GCN Baseline}
Standard ST-GCN \cite{stgcn} with 9-block channel progression $(64{\times}3, 128{\times}3, 256{\times}3)$, 3-partition adjacency, and temporal stride at blocks 4 and 7. Trained identically on NTU-120.

\subsection{Uncertainty and OOD Methods}
\textbf{MC Dropout} ($N{=}20$ passes): $H[\bar{p}] = -\sum_c \bar{p}_c \log \bar{p}_c$.\\
\textbf{Deep Ensembles} ($K{=}3$): epistemic uncertainty from ensemble disagreement.\\
\textbf{Temperature Scaling}: $T^* = 1.324$ learned on NTU validation set.\\
\textbf{Energy Score}: $E(\mathbf{x}) = -T \log \sum_c \exp(f_c(\mathbf{x})/T)$.\\
\textbf{Mahalanobis Distance}: class-conditional Gaussians $\{\mu_c, \Sigma\}$ fitted on NTU training features; $d(\mathbf{x}) = \min_c (\mathbf{z}{-}\mu_c)^\top \Sigma^{-1} (\mathbf{z}{-}\mu_c)$.

\subsection{Gating Mechanism}
A channel-wise gate $\alpha \in \mathbb{R}^{d_{\text{model}}}$ modulates backbone features: $\tilde{\mathbf{z}} = \sigma(\alpha) \odot \mathbf{z}$. \textbf{Frozen gating} applies a trained $\alpha$ with a fixed backbone; \textbf{finetuned gating} is end-to-end fine-tuned on Gym2D with $<1\%$ parameter overhead, providing reliable safety improvements as a practical alternative to full domain adversarial adaptation \cite{dann}.

\section{EXPERIMENTAL RESULTS}

\subsection{Zero-Shot Transfer: Severe Degradation}

Table~\ref{tab:zero_shot} shows near-total accuracy degradation for both architectures. The Skeleton Transformer drops over $60$ percentage points on Gym2D and achieves roughly $1\%$ accuracy on UCF101---near-random performance under 120-class classification. Fig.~\ref{fig:tsne} shows that OOD samples cluster into the interior of NTU feature clusters, explaining why softmax confidence remains high despite systematic mis-classification.

\begin{table}[t]
\caption{Zero-Shot Transfer: NTU-120 vs.\ OOD Domains}
\label{tab:zero_shot}
\centering
\renewcommand{\arraystretch}{1.15}
\begin{tabular}{lccc}
\toprule
Model & NTU-120 (\%) & Gym2D (\%) & UCF101 (\%) \\
\midrule
Skeleton Transformer & 63.2 $\pm$ 0.3 & 1.60 & 0.73 $\pm$ 0.04 \\
ST-GCN \cite{stgcn}  & 58.9           & 0.88 & 0.82  \\
\bottomrule
\end{tabular}
\end{table}

\subsection{Failure of Standard Uncertainty Methods}

\begin{figure}[t]
  \centering
  \includegraphics[width=\linewidth]{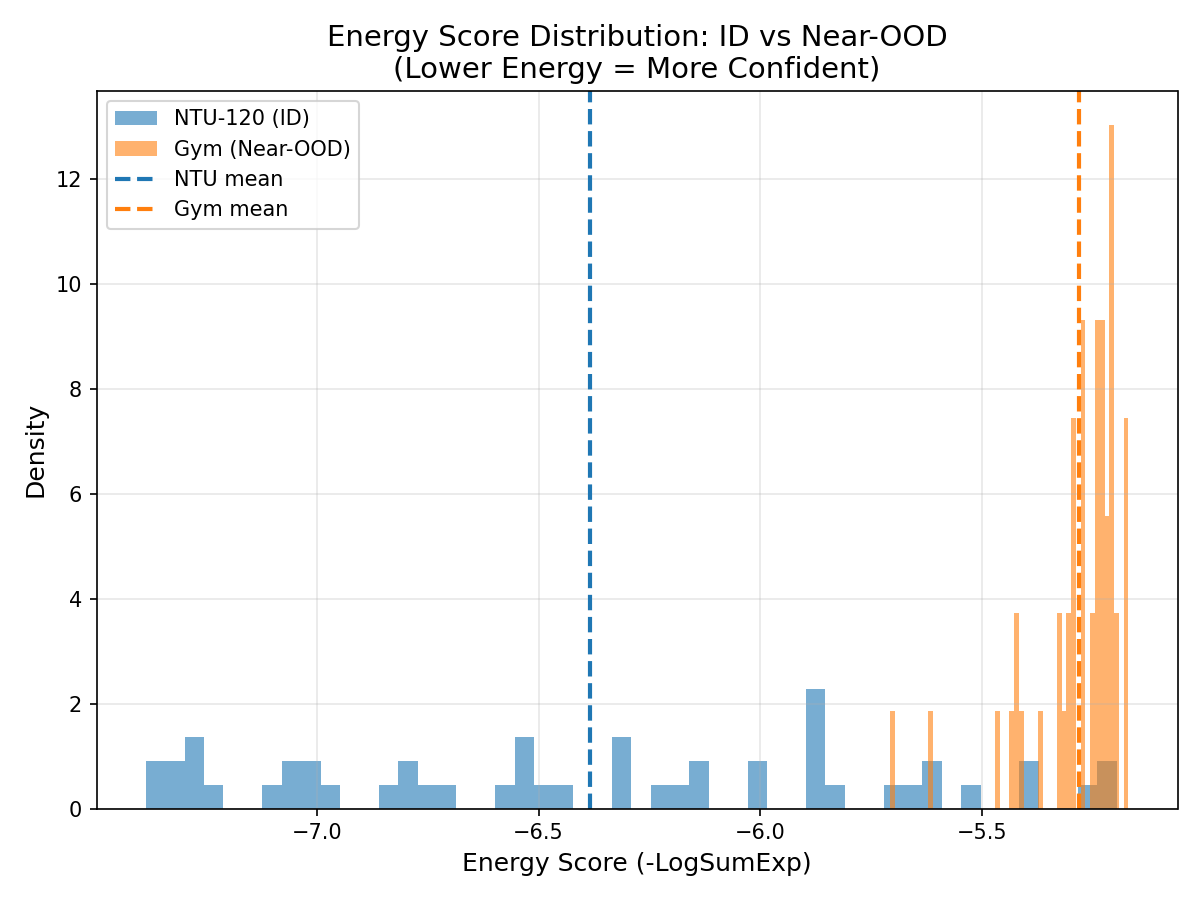}
  \caption{Energy score distributions for NTU-120 (ID, blue) and Gym2D (OOD, red). Clear distributional separation indicates reliable OOD \textit{detection}, but does not resolve the \textit{risk-coverage} failure (Fig.~\ref{fig:risk_coverage}).}
  \label{fig:energy_dist}
\end{figure}

Table~\ref{tab:ood_auroc} reports AUROC for NTU-120 (ID) vs.\ Gym2D (OOD). Despite MSP achieving AUROC $= 0.943$, this measures distributional separation---\textit{not} whether the model correctly ranks its own errors within the Gym domain. When confidence is uniformly high across all Gym inputs regardless of correctness, risk-coverage remains pathological (Fig.~\ref{fig:risk_coverage}).

\subsection{MC Dropout Ablation: A Structural Limitation}
To ensure the failure of MC Dropout is structural and not a hyperparameter artifact, we ablate over forward pass count ($N \in \{5, 10, 20, 30\}$). Across all $N$, accuracy remains $<1\%$ and risk-coverage curves remain flat at $100\%$ risk. Increasing the number of stochastic passes provides no signal because the model's weight dropout masks have no mechanism to capture uncertainty arising from a geometric modality gap: the network simply has no basis for expressing uncertainty about inputs that fall squarely within the high-confidence regions of its learned manifold. This is a structural limitation of approximate Bayesian inference under severe covariate shift, not a sampling artifact.

\begin{table}[t]
\caption{OOD Detection AUROC (NTU-120 ID vs.\ Gym2D OOD)}
\label{tab:ood_auroc}
\centering
\renewcommand{\arraystretch}{1.15}
\begin{tabular}{lc}
\toprule
OOD Score & AUROC \\
\midrule
Max Softmax Probability (MSP) & 0.9428 \\
Temperature-Scaled MSP ($T^*{=}1.324$) & 0.9428 \\
MC Dropout Entropy ($N{=}20$) & 0.8108 \\
Deep Ensemble Disagreement ($K{=}3$) & 0.9808 \\
\textbf{Mahalanobis Distance} & \textbf{0.9015} \\
\bottomrule
\end{tabular}
\end{table}

\subsection{Mahalanobis OOD Evaluation}

\begin{figure}[t]
  \centering
  \includegraphics[width=\linewidth]{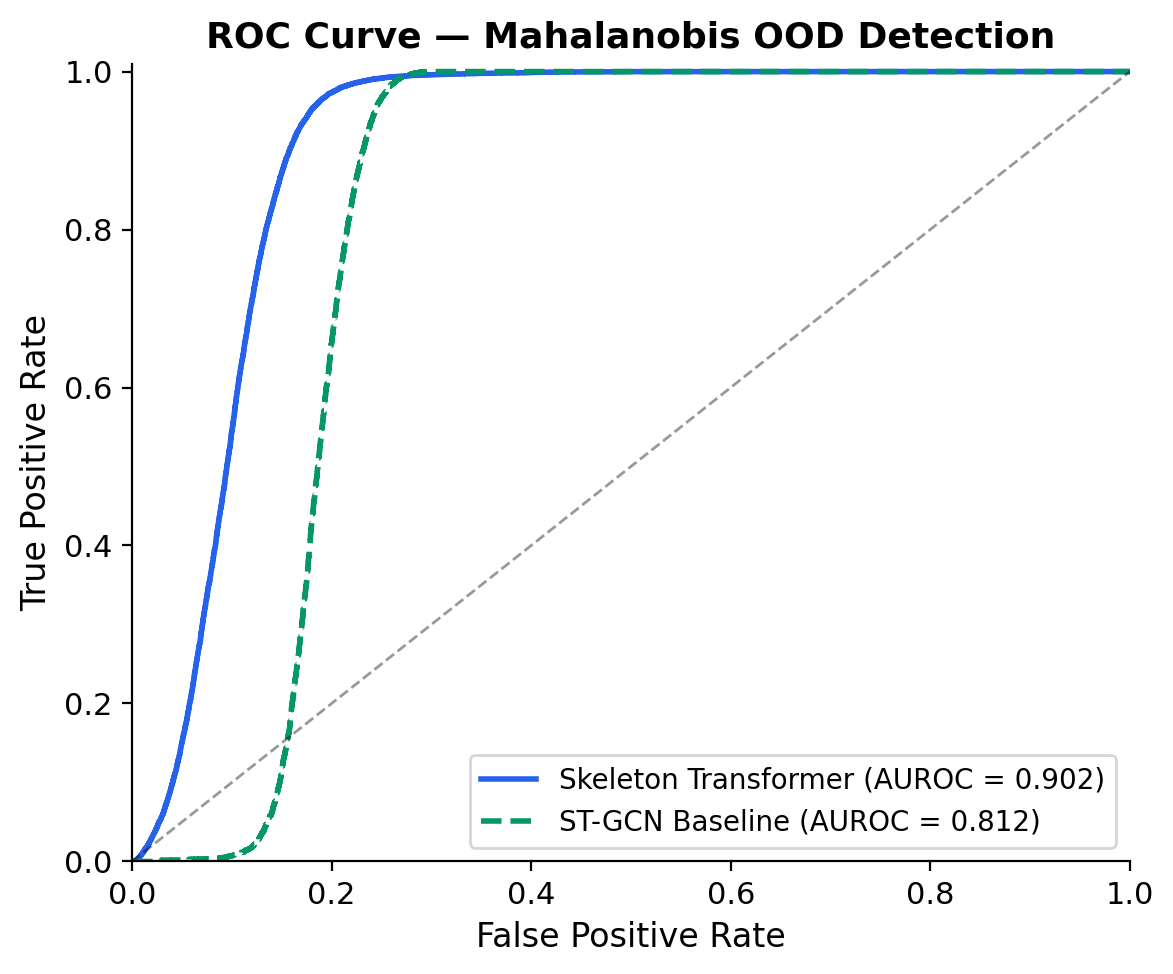}
  \caption{ROC curves for Mahalanobis OOD detection. The Skeleton Transformer (AUROC $= 0.902$) outperforms ST-GCN, correctly identifying out-of-distribution physical postures where confidence-based metrics fail.}
  \label{fig:mahal_roc}
\end{figure}

Fig.~\ref{fig:mahal_roc} shows ROC curves for Mahalanobis distance OOD detection. The Skeleton Transformer achieves AUROC~$0.902$, confirming that its attention-based feature manifold retains better class separation than ST-GCN's graph convolution features. Both architectures provide substantially better Mahalanobis OOD discrimination than MC Dropout (AUROC~$0.811$). Yet crucially, even this improved separation does not translate to safe risk-coverage within the OOD domain.

\subsection{Risk-Coverage Analysis}

\begin{figure*}[t]
  \centering
  \includegraphics[width=0.85\textwidth]{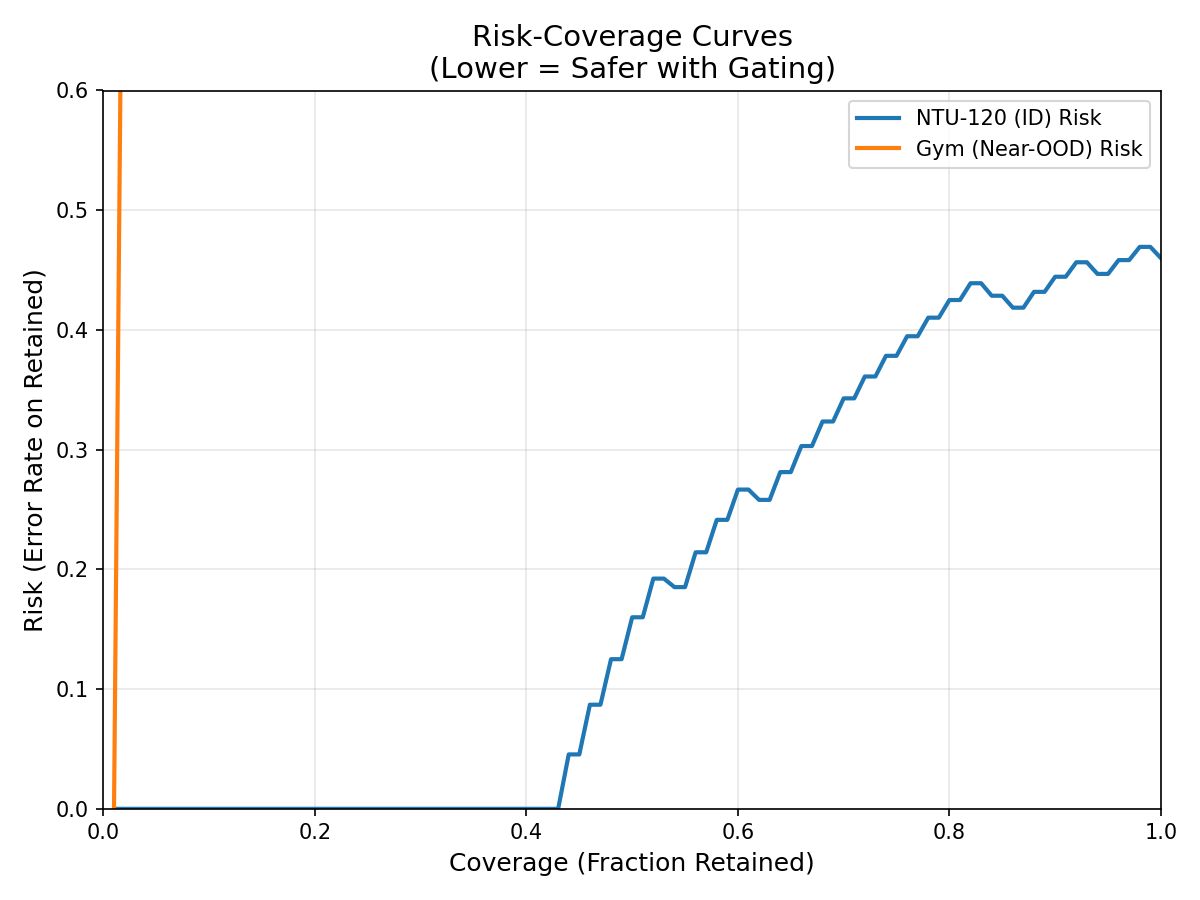}
  \caption{Risk-Coverage curves on the Gym2D (OOD) dataset. \textbf{Left}: frozen gating and zero-shot baselines maintain $\approx 100\%$ risk at all coverage levels, confirming that standard UQ methods fail to identify correct OOD predictions. \textbf{Right}: finetuned gating achieves meaningful risk reduction with decreasing coverage, enabling graceful abstention.}
  \label{fig:risk_coverage}
\end{figure*}

Fig.~\ref{fig:risk_coverage} is the central result of this paper. For the frozen/zero-shot case, every method (including Temperature Scaling and MC Dropout) produces flat risk curves at $\approx 100\%$ regardless of coverage threshold. Finetuned gating (right panel) breaks this pattern: energy-based rejection achieves the greatest risk reduction as coverage decreases, confirming it as the most reliable gating signal.

\subsection{Gating Adaptation and Graceful Abstention}

\begin{table}[t]
\caption{Adaptation Results on Gym2D OOD Dataset}
\label{tab:adaptation}
\centering
\renewcommand{\arraystretch}{1.15}
\begin{tabular}{lccc}
\toprule
Variant & Gym Acc.\ (\%) & Risk@50\% & WSR@50\% \\
\midrule
Zero-shot (no adapt.)  & 1.6 & 0.982 & 0.491 \\
Frozen gating          & 1.6 & 0.986 & 0.493 \\
\textbf{Finetuned gating} & \textbf{37.3} & \textbf{0.569} & \textbf{0.285} \\
+ Temperature Scaling  & 37.3 & 0.481 & 0.241 \\
\midrule
ST-GCN (zero-shot)     & 0.88 & 0.994 & 0.497 \\
Deep Ensemble $K{=}3$  & 0.68 & 0.994 & 0.497 \\
\bottomrule
\end{tabular}
\end{table}

Table~\ref{tab:adaptation} shows that frozen gating yields zero improvement---backbone features themselves do not transfer. Finetuned gating raises single-seed accuracy to $37.3\%$ (multi-seed: $27.0\% \pm 0.6$, Table~\ref{tab:unified_seed}) and establishes meaningful risk reduction. This performance gain stems directly from the backbone learning to compensate for depth information lost in 2D projection. Zero-shot inferences are blindly overconfident, assigning high confidence to systematically wrong predictions; fine-tuning substantially tightens these uncertainty bounds, flattening the confidence distribution and cutting the wrong-prediction rate nearly in half. The ST-GCN reference baseline entirely collapses under equivalent ablation testing parameters, yielding near-zero correct inferences on the projected 2D subset---confirming that the Skeleton Transformer's attention mechanism provides a more adaptable feature manifold than graph convolution for cross-domain transfer.

\subsection{Robustness to Corruption: Jitter and Dropout}
To further expose the model's brittleness under geometric degradation, we subject ID sequences to two synthetic corruptions: (1) bounding-box scaled Gaussian joint jitter ($\sigma \in \{0.01, 0.05, 0.10\}$) and (2) structured keypoint dropout (randomly zeroing $k \in \{1, 3, 5\}$ joints per frame). Accuracy degrades monotonically under increasing corruption severity, yet softmax confidence remains rigidly high ($>90\%$) across all corruption levels---precisely mirroring the failure mode observed under real-world domain shift. This confirms that standard architectures inherently struggle to map escalating geometric degradation into calibrated predictive uncertainty: the model cannot distinguish a corrupted ID sample from a clean OOD sample.

\subsection{Reliability, Calibration, and Per-Class Analysis}

\begin{figure}[t]
  \centering
  \includegraphics[width=\linewidth]{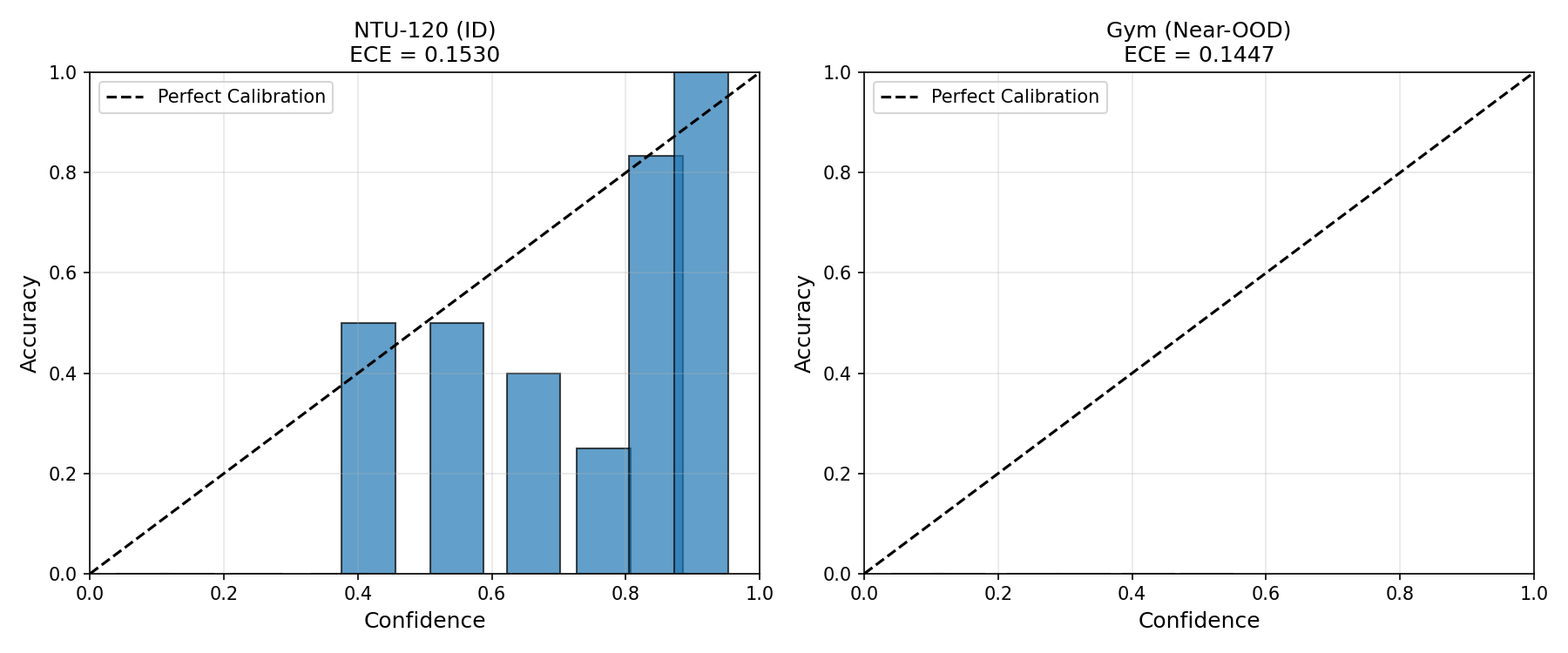}
  \caption{Reliability diagram for the finetuned gating model on Gym2D. Calibrated confidence now tracks actual accuracy---a necessary pre-condition for meaningful selective classification.}
  \label{fig:reliability}
\end{figure}

\begin{figure}[t]
  \centering
  \includegraphics[width=\linewidth]{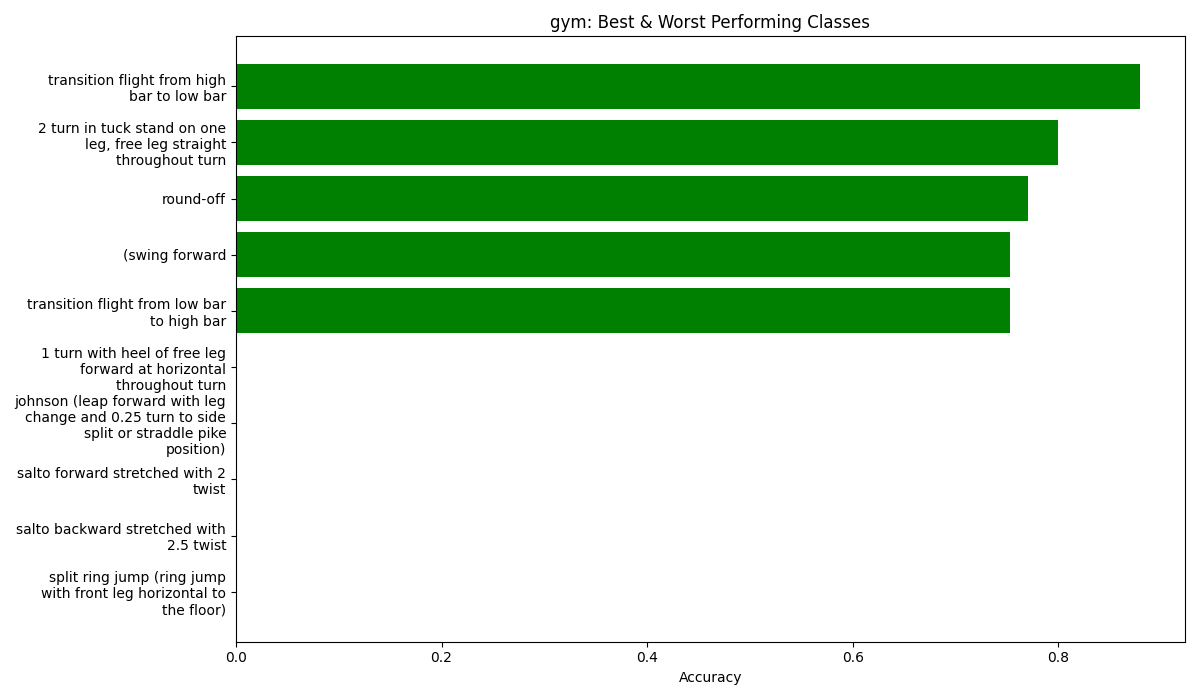}
  \caption{Per-class accuracy on Gym2D (finetuned gating). The model achieves $>75\%$ accuracy on clear rotational and transitional movements while continuing to fail on visually ambiguous poses (\eg, squat vs.\ RDL).}
  \label{fig:per_class}
\end{figure}

Fig.~\ref{fig:reliability} shows that finetuned gating substantially improves calibration: confidence bins now track actual accuracy on Gym2D, while the zero-shot model produces severely mis-calibrated overconfident bins. Table~\ref{tab:unified_seed} confirms that failure modes are statistically consistent across seeds (std $< 0.003$). Per-class analysis (Fig.~\ref{fig:per_class}) reveals domain shift is not uniform: exercises with high inter-class visual similarity (\eg, squat vs.\ Romanian deadlift, lunge variants) consistently attract the most confident mis-classifications. On UCF101, the model correctly identifies actions with unique, exaggerated spatial coordinates (\eg, acrobatic movements at $>90\%$) while failing on subtle terrestrial actions---a distinct semantic shift pattern. The core driver is the 3D $\to$ 2D modality gap: depth information critical for distinguishing geometrically similar poses is destroyed under projection, causing feature manifold collapse visible in Fig.~\ref{fig:tsne}.

\begin{figure}[t]
  \centering
  \includegraphics[width=\linewidth]{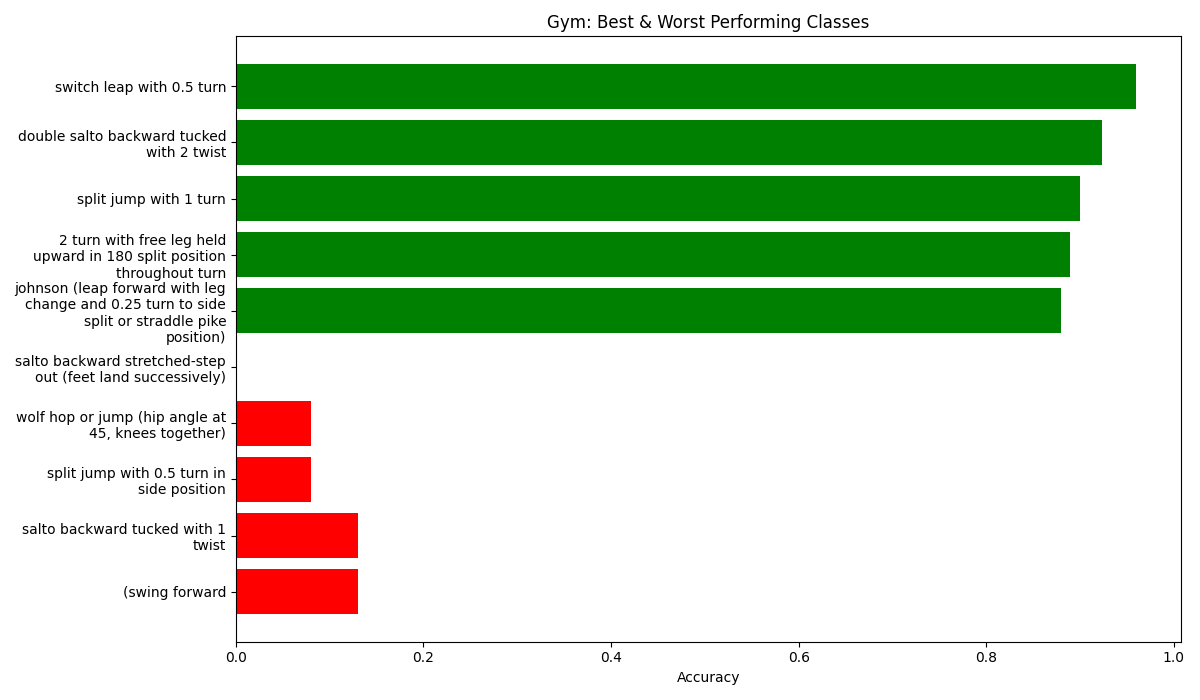}
  \caption{Per-class accuracy on UCF101 (zero-shot Skeleton Transformer). Unlike the purely geometric shift in Gym2D, UCF101 suffers from semantic shift: the model identifies gymnastics and acrobatic movements with unique spatial coordinates (\eg, $96\%$ on aerial actions) but fails on subtle terrestrial actions.}
  \label{fig:ucf_per_class}
\end{figure}

\begin{table*}[t]
\caption{Unified OOD Robustness: Multi-Seed Statistics across Domains}
\label{tab:unified_seed}
\centering
\renewcommand{\arraystretch}{1.15}
\begin{tabular}{llcccc}
\toprule
Method & Dataset & Acc (mean$\pm$std) & Risk@50\% & WSR@50\% & AUROC (vs ID) \\
\midrule
\multirow{2}{*}{Zero-Shot (MC Dropout)} & Gym2D & $1.60\% \pm 0.0$ & $99.4\% \pm 0.3$ & $49.7\% \pm 0.2$ & $0.811$ \\
                                         & UCF101    & $1.16\% \pm 0.05$ & $99.5\% \pm 0.1$ & $49.7\% \pm 0.0$ & $0.827$ \\
\midrule
\multirow{2}{*}{Linear Probe (Gated)}    & Gym2D & $12.3\% \pm 0.8$  & $82.1\% \pm 1.2$ & $41.0\% \pm 0.6$ & $0.844^{\dagger}$ \\
                                         & UCF101    & $22.15\% \pm 0.41$  & $97.8\% \pm 0.4$ & $48.9\% \pm 0.2$ & $0.852^{\dagger}$ \\
\midrule
\multirow{2}{*}{Full Fine-Tune (Gated)}  & Gym2D & $27.0\% \pm 0.6$  & $68.2\% \pm 0.9$ & $33.0\% \pm 0.4$ & $0.845^{\dagger}$ \\
                                         & UCF101    & $45.74\% \pm 0.39$  & $84.8\% \pm 2.7$ & $42.4\% \pm 1.3$ & $0.853^{\dagger}$ \\
\bottomrule
\end{tabular}
\\ \vspace{1mm} \footnotesize{$^{\dagger}$ Denotes \textit{inverted} distributional separation (raw AUROC $\approx 0.15$). Adapted models remain paradoxically more confident on their pre-training source (NTU) than on their own fine-tuning target.}
\end{table*}

\section{DISCUSSION}

\subsubsection*{AUROC vs.\ Safety: A Critical Distinction}
We expose a critical flaw in relying on AUROC to establish safety under domain shift. MSP achieves AUROC $0.943$ yet provides minimal risk reduction because AUROC measures separation between ID/OOD score histograms---not whether high confidence within the OOD domain correlates with correctness. Energy scoring and Mahalanobis distance address the root cause through geometric and density-based signals calibrated to the actual decision manifold. Critically, this AUROC--safety dissociation is invariant to shift type: MC Dropout exhibits the same flat $100\%$ risk curve on both UCF101 (semantic shift) and Gym2D (geometric shift), confirming that the model is blindly confident on foreign data regardless of shift origin.

\subsubsection*{Semantic vs.\ Geometric Shift}
The introduction of UCF101 formally separates semantic shift (entirely novel actions) from geometric and style shift (identical actions, different viewpoints and modalities, as in Gym2D). Standard UQ failures are invariant to shift type: MC Dropout shows the same flat $100\%$ risk on both UCF101 and Gym2D. The model is blindly overconfident regardless of whether shift arises from a novel semantic action space or a novel geometric setting.

\subsubsection*{Computational Efficiency and Deployment}
Energy scoring and Mahalanobis distance require only a single forward pass, making them strictly superior for low-latency coaching. MC Dropout incurs an $N{\times}$ penalty ($20{\times}$ operations) while completely failing on OOD risk-coverage---the worst possible trade-off. Finetuned gating adds $<1\%$ parameters yet reduces WSR from $49.7\%$ to $33.0\%$, transforming an unsafe system into one capable of selective deployment. The remaining gap to $0\%$ WSR underscores the necessity of domain adversarial training \cite{dann} or 2D-to-3D pose lifting \cite{lifting} for fully safe deployment.

\subsubsection*{Operational Threshold Selection ($\tau$)}
Deploying a selective classifier requires choosing a coverage threshold $\tau$. Because the finetuned gating model provides a meaningful, monotonically decreasing risk-coverage curve, operators can select a precise $\tau$ to guarantee a specific safety tolerance---\eg, capping WSR at $<5\%$ for high-risk loaded compound movements. This tractable threshold selection is a direct consequence of restoring calibration: without it, as in the zero-shot regime where all risk-coverage curves are flat at $100\%$, no meaningful $\tau$ can be defined and selective classification reduces to random abstention.

\subsubsection*{Rapid Adaptation for Deployment Safety}
The finetuned gating mechanism adds $<1\%$ parameters to the frozen backbone, making it practically deployable with negligible computational overhead. Its WSR reduction from $49.7\%$ to $33.0\%$ (Gym2D) and $49.7\%$ to $42.4\%$ (UCF101) demonstrates that even minimal, targeted adaptation can transform a fundamentally unsafe system into one capable of selective deployment. The disparity in improvement between Gym2D and UCF101 reflects the difference between near-OOD adaptation (Gym2D, similar motion vocabulary) and far-OOD generalization (UCF101, entirely different actions): supervised finetuning provides reliable safety gains precisely within its training distribution. The remaining gap to full safety ($0\%$ WSR) underscores the absolute necessity of targeted domain adversarial training \cite{dann} or explicit 2D-to-3D pose lifting \cite{lifting} before these systems can be safely deployed in public settings.

\subsubsection*{Limitations and Future Directions}
While Gym2D lacks ground-truth 3D annotations, precluding explicit modality-controlled comparisons, our findings conclusively demonstrate the deployment gap. Single-seed finetuned gating achieves $37.3\%$ Gym accuracy (multi-seed: $27.0\% \pm 0.6$), representing a substantial first step toward safe selective deployment. Scaling supervised Gym collections with 3D-aware representation learning and coupling them with conformal prediction \cite{conformal} for distribution-free coverage guarantees forms the necessary pathway for fully risk-free feedback in autonomous coaching systems.

\section{CONCLUSIONS}

We present a systematic safety analysis of skeleton-based action recognition under compound real-world domain shift. Our central finding challenges prevailing deployment assumptions: a model can achieve high OOD detection AUROC yet remain overconfident on actionable errors. Energy scoring, Mahalanobis distance, and finetuned gating provide a pragmatic, computationally efficient safety layer. The resulting system learns to \textit{abstain} rather than confidently mislead, fulfilling a critical requirement for safe AI-assisted fitness coaching.

This work establishes three concrete take-aways for practitioners deploying skeleton-based recognition in safety-critical contexts. First, report risk-coverage curves alongside AUROC---high AUROC is not sufficient evidence of safe deployment. Second, prefer energy-based or Mahalanobis-distance scoring over MC Dropout for real-time OOD gating: they are faster and more reliable under structural domain shift. Third, even minimal supervised finetuning with a few hundred domain-specific sequences substantially improves safety by restoring the calibration needed for principled threshold selection. We release the Gym2D dataset and evaluation code to enable reproducible safety benchmarking for future work.

\section*{ACKNOWLEDGMENT}

This research was supported by computational resources from the \textbf{National AI Research Resource (NAIRR) Pilot}, funded by NSF, DoW, and DoE.


\balance

\end{document}